\documentclass{article}

\usepackage{arxiv}
\usepackage[utf8]{inputenc}
\usepackage[T1]{fontenc}
\usepackage{hyperref}
\usepackage{url}
\usepackage{booktabs}
\usepackage{amsfonts}
\usepackage{amsmath}
\usepackage{nicefrac}
\usepackage{microtype}
\usepackage{cleveref}
\usepackage{graphicx}
\usepackage{natbib}
\usepackage{doi}
\usepackage{multirow}
\usepackage{array}
\usepackage{subcaption}
\usepackage{listings}
\usepackage{xcolor}
\usepackage{tabularx}
\usepackage{float}
\usepackage{needspace}

\renewcommand{\headeright}{}
\renewcommand{\undertitle}{}
\title{A Comparative Analysis of Classical Machine Learning and Deep Learning Approaches for Sentiment Classification on IMDb Movie Reviews}

\author{%
  Erma Daniar Safitri\\
  Department of Data Science\\
  Institut Teknologi Sumatera\\
  Lampung Selatan, 35365, Indonesia\\
  \texttt{erma.123450061@student.itera.ac.id}
  \And
  Lia Hana Ichisasmita\\
  Department of Data Science\\
  Institut Teknologi Sumatera\\
  Lampung Selatan, 35365, Indonesia\\
  \texttt{lia.123450089@student.itera.ac.id}
  \AND
  Citra Agustin\\
  Department of Data Science\\
  Institut Teknologi Sumatera\\
  Lampung Selatan, 35365, Indonesia\\
  \texttt{citra.123450108@student.itera.ac.id}
  \And
  Luluk Muthoharoh, S.Si., M.Si.\\
  Department of Data Science\\
  Institut Teknologi Sumatera\\
  Lampung Selatan, 35365, Indonesia\\
  \texttt{luluk.muthoharoh@sd.itera.ac.id}\\
  \AND
  Ardika Satria, S.Si., M.Si.\\
  Department of Data Science\\
  Institut Teknologi Sumatera\\
  Lampung Selatan, 35365, Indonesia\\
  \texttt{ardika.satria@sd.itera.ac.id}
  \And
  Martin Clinton Tosima Manullang, Ph.D.\\
  Department of Informatics Engineering\\
  Institut Teknologi Sumatera\\
  Lampung Selatan, 35365, Indonesia\\
  \texttt{martin.manullang@if.itera.ac.id}\\
}

\renewcommand{\shorttitle}{A Comparative Analysis of Classical Machine Learning and Deep Learning Approaches for Sentiment Classification on IMDb Movie Reviews}

\hypersetup{
    pdftitle = {A Comparative Analysis of Classical Machine Learning and Deep Learning Approaches for Sentiment Classification on IMDb Movie Reviews},
    pdfsubject = {cs.CL, cs.LG},
    pdfauthor = {Erma Daniar Safitri, Lia Hana Ichiasmita, Citra Agustin, Martin Clinton Tosima Manullang},
    pdfkeywords = {Sentiment Analysis, IMDb Dataset, Machine Learning, Deep Learning, BiLSTM, Attention Mechanism, PyCaret, TF-IDF, Text Classification, pba2026-kelompok-3}
}

\begin{document}
\maketitle

\begin{abstract}
Sentiment analysis is a fundamental task in Natural Language Processing (NLP) that aims to determine the polarity of textual data. This study presents a comparative analysis between classical machine learning (ML) and deep learning (DL) approaches for sentiment classification using the IMDb movie reviews dataset. The ML pipeline utilizes PyCaret AutoML with TF-IDF feature extraction to evaluate models such as Logistic Regression, Naïve Bayes, and Support Vector Machine (SVM). Meanwhile, the DL pipeline employs sequence-based models, including Bidirectional Long Short-Term Memory (BiLSTM) and BiLSTM with an attention mechanism.

Experimental results show that classical ML models, particularly SVM, achieve superior performance with an accuracy of 0.8530, outperforming DL models in this study. However, the BiLSTM with Attention model demonstrates improved contextual understanding compared to the standard BiLSTM, achieving an accuracy of 0.706. These findings highlight that while deep learning models are capable of capturing sequential dependencies, traditional machine learning approaches remain strong baselines when combined with effective feature engineering such as TF-IDF.

This study provides insights into the trade-offs between performance and computational efficiency in sentiment classification tasks, emphasizing that model selection should consider both dataset size and available resources. The source code is publicly available at: \url{https://github.com/Ermadaniarsafitri061/pba2026-kelompok-3}
\end{abstract}


\section{Introduction}
\label{sec:intro}

The rapid growth of user-generated content on online platforms has significantly increased the importance of sentiment analysis in Natural Language Processing (NLP). Movie review platforms such as IMDb provide large-scale opinionated text data that can be utilized to understand audience perception and support decision-making processes. These reviews play an important role in influencing user decisions and reflect public opinion toward movies \cite{ramadhan2022}.

Sentiment analysis aims to classify textual data into predefined categories such as positive or negative sentiment. Traditional machine learning (ML) approaches such as Logistic Regression, Naïve Bayes, and Support Vector Machine (SVM) have been widely used due to their efficiency and effectiveness in text classification tasks \cite{subedi2025}. These methods typically rely on feature extraction techniques such as TF-IDF to convert textual data into numerical representations \cite{yasen2019}.

In contrast, deep learning (DL) approaches have gained significant attention due to their ability to automatically learn feature representations from raw text data. Models such as Long Short-Term Memory (LSTM) and Bidirectional LSTM (BiLSTM) are capable of capturing contextual and sequential dependencies in text, leading to improved classification performance \cite{dang2020}. Furthermore, recent studies show that BiLSTM-based models outperform traditional ML approaches in sentiment classification tasks \cite{singh2023}.

Despite their superior performance, deep learning models require higher computational resources and longer training time compared to classical ML methods. Therefore, it is important to conduct a comparative analysis to understand the trade-offs between these approaches in terms of performance, efficiency, and practicality.

Using the IMDb movie reviews dataset obtained from Kaggle, this study proposes a comparative framework between classical machine learning and deep learning approaches for sentiment classification.

This work makes the following contributions:

\begin{enumerate}
    \item A dual preprocessing pipeline for sentiment analysis, consisting of:
    \begin{itemize}
        \item Machine Learning preprocessing using an EDA-based pipeline implemented in a Jupyter Notebook, including text normalization and TF-IDF feature extraction.
        \item Deep Learning preprocessing implemented in \texttt{preprocess.py}, including text cleaning (lowercasing, URL removal, HTML removal, non-alphabet filtering), stratified sampling of 10,000 data, and label encoding.
    \end{itemize}

    \item A machine learning model built using PyCaret AutoML with TF-IDF features, serialized as \texttt{sentiment\_model.pkl} for efficient inference and deployment.

    \item A deep learning framework consisting of two architectures:
    \begin{itemize}
        \item BiLSTM model for sequential text modeling
        \item BiLSTM with Attention mechanism to enhance contextual representation
    \end{itemize}

    \item An interactive deployment of both ML and DL models through Hugging Face Spaces:
    
    \url{https://huggingface.co/spaces/Ermadaniarsafitri061/imdb-sentiment}
\end{enumerate}

\section{Related Work}
\label{sec:related}

\subsection{Machine Learning for Sentiment Analysis}

Sentiment analysis has traditionally been addressed using classical machine learning approaches. Algorithms such as Naïve Bayes, Logistic Regression, and Support Vector Machine (SVM) have been widely applied due to their simplicity and efficiency. These models typically rely on feature extraction techniques such as bag-of-words and Term Frequency--Inverse Document Frequency (TF-IDF) to convert textual data into numerical representations \cite{subedi2025,yasen2019}.

Several studies have demonstrated the effectiveness of machine learning models on sentiment classification tasks using IMDb datasets. In particular, SVM-based approaches have shown strong performance due to their ability to handle high-dimensional sparse data \cite{ramadhan2022}. However, these methods depend heavily on feature engineering and are limited in capturing contextual relationships between words.

\subsection{Deep Learning for Sentiment Analysis}

With the advancement of deep learning, neural network-based approaches have become increasingly popular for sentiment analysis. Models such as Recurrent Neural Networks (RNN) and Long Short-Term Memory (LSTM) are capable of modeling sequential dependencies in text, enabling better understanding of context \cite{dang2020}.

Bidirectional LSTM (BiLSTM) extends this capability by processing text in both forward and backward directions, which improves contextual representation and classification performance \cite{singh2023}. Recent studies have shown that deep learning models often outperform traditional machine learning approaches in sentiment analysis tasks \cite{gupta2024,walji2025}.

Despite their advantages, deep learning models require higher computational resources and longer training time, which may limit their practical applicability in certain scenarios.

\subsection{Comparison between Machine Learning and Deep Learning}

Although deep learning approaches generally achieve higher accuracy, classical machine learning methods remain relevant due to their efficiency, interpretability, and lower computational cost. Machine learning models are particularly suitable for scenarios with limited computational resources or smaller datasets.

Therefore, comparative studies between machine learning and deep learning approaches are essential to understand their respective strengths and limitations. Such comparisons provide valuable insights for selecting appropriate models based on specific application requirements, including performance, scalability, and resource constraints.mark, alongside classical machine-learning baselines.

\section{Dataset}
\label{sec:dataset}

This study uses the IMDb movie reviews dataset obtained from Kaggle. The dataset is obtained from Kaggle \url{https://www.kaggle.com/datasets/lakshmi25npathi/imdb-dataset-of-50k-movie-reviews}. The dataset is widely used as a benchmark in sentiment analysis research due to its large size and balanced class distribution \cite{singh2023,yasen2019}.

\subsection{Data Description}

Each data instance consists of a textual movie review and its corresponding sentiment label. The dataset contains 50,000 samples divided into two classes: positive and negative, with an equal distribution of 25,000 samples per class. This balanced distribution helps reduce bias during model training and evaluation \cite{dong2020}.

\subsection{Dataset Statistics}

Table~\ref{tab:stats} presents the distribution of sentiment labels in the dataset. The balanced nature of the dataset ensures fair evaluation of classification models without class imbalance issues.

\begin{table}[h]
\centering
\begin{tabular}{lc}
\toprule
\textbf{Sentiment} & \textbf{Number of Samples} \\
\midrule
Positive & 25,000 \\
Negative & 25,000 \\
\midrule
\textbf{Total} & \textbf{50,000} \\
\bottomrule
\end{tabular}
\caption{Dataset statistics}
\label{tab:stats}
\end{table}

\subsection{EDA and Preprocessing}

\begin{figure}[H]
\centering
\includegraphics[width=0.3\linewidth]{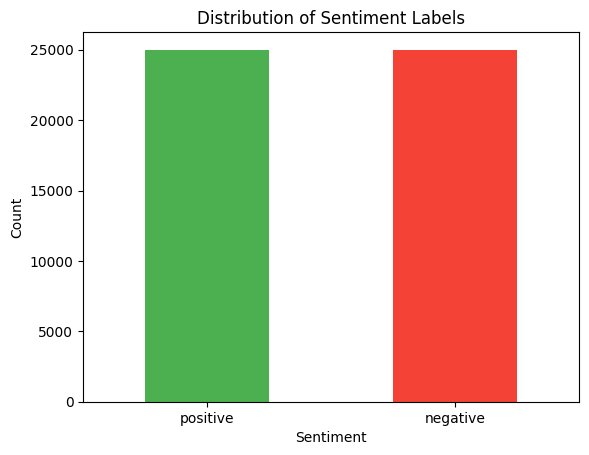}
\caption{Distribution of sentiment labels}
\label{fig:eda}
\end{figure}

Exploratory Data Analysis (EDA) is conducted to understand the dataset characteristics prior to model development. As shown in Figure~\ref{fig:eda}, the dataset is evenly distributed across sentiment classes, which helps improve model reliability.

Basic preprocessing steps include text normalization such as converting text to lowercase, removing noise (e.g., punctuation, URLs, and HTML tags), and whitespace normalization. These steps are essential to improve data quality and reduce irrelevant information \cite{fitroh2023}.

For the machine learning (ML) pipeline, the cleaned text is transformed into numerical features using TF-IDF, which is effective for traditional text classification models \cite{tarimer2024}. For the deep learning (DL) pipeline, additional steps such as stratified sampling, label encoding, and sequence padding are applied to prepare the data for neural network models.

\section{Methodology}

This study adopts a comparative framework to evaluate classical machine learning (ML) and deep learning (DL) approaches for sentiment classification. The overall pipeline consists of data preprocessing, feature representation, model training, and evaluation.

\subsection{Machine Learning Approach}

For the machine learning pipeline, textual data is transformed into numerical features using Term Frequency--Inverse Document Frequency (TF-IDF), which is widely used in text classification tasks \cite{manning2008,tarimer2024}.

TF-IDF is defined as:

\begin{equation}
TF\text{-}IDF(t,d) = TF(t,d) \times IDF(t)
\end{equation}

where:

\begin{equation}
TF(t,d) = \frac{f_{t,d}}{\sum_{t'} f_{t',d}}
\end{equation}

\begin{equation}
IDF(t) = \log \frac{N}{df(t)}
\end{equation}

Here, $f_{t,d}$ represents the frequency of term $t$ in document $d$, $N$ is the total number of documents, and $df(t)$ is the number of documents containing term $t$.

Several machine learning models are evaluated, including Logistic Regression, Naïve Bayes, and Support Vector Machine (SVM), which are commonly used for sentiment classification \cite{ramadhan2022,subedi2025}.

\subsection{Deep Learning Approach}

For the deep learning pipeline, sequence-based models are used to capture contextual information in text. Long Short-Term Memory (LSTM) networks are widely used due to their ability to model long-term dependencies in sequential data \cite{hochreiter1997,dang2020}.

The LSTM unit is defined as follows:

\begin{align}
f_t &= \sigma(W_f x_t + U_f h_{t-1} + b_f) \\
i_t &= \sigma(W_i x_t + U_i h_{t-1} + b_i) \\
o_t &= \sigma(W_o x_t + U_o h_{t-1} + b_o) \\
\tilde{c}_t &= \tanh(W_c x_t + U_c h_{t-1} + b_c) \\
c_t &= f_t \odot c_{t-1} + i_t \odot \tilde{c}_t \\
h_t &= o_t \odot \tanh(c_t)
\end{align}

where $f_t$, $i_t$, and $o_t$ denote the forget, input, and output gates, respectively.

To enhance performance, a Bidirectional LSTM (BiLSTM) is used, which processes sequences in both forward and backward directions \cite{schuster1997}.

Additionally, an attention mechanism is incorporated to improve the model’s ability to focus on important words within the sequence \cite{bahdanau2015,gupta2024}. The attention mechanism is defined as:

\begin{equation}
\alpha_t = \frac{\exp(e_t)}{\sum_{k} \exp(e_k)}
\end{equation}

\begin{equation}
c = \sum_{t} \alpha_t h_t
\end{equation}

where $\alpha_t$ represents the attention weights and $h_t$ denotes hidden states.

\subsection{Model Comparison Strategy}

To ensure a fair comparison, both ML and DL models are evaluated using the same dataset and evaluation metrics. ML models rely on feature-based representations such as TF-IDF, while DL models learn contextual representations directly from sequences.

This comparison allows for analyzing the trade-offs between classical and deep learning approaches in terms of performance and computational efficiency \cite{walji2025}.

\section{Experiments}

This section describes the experimental configuration, hyperparameter settings, and evaluation metrics used to compare machine learning (ML) and deep learning (DL) approaches.

\subsection{Experimental Configuration}

For the ML pipeline, PyCaret AutoML is used to train and compare multiple models. The dataset is split into training and testing sets using an 80:20 ratio with a fixed random seed (42) to ensure reproducibility. TF-IDF is applied automatically as part of the text feature extraction process.

For the DL pipeline, experiments are conducted using PyTorch on a subset of 10,000 samples selected via stratified sampling. The dataset is divided into training, validation, and test sets with proportions of 70\%, 10\%, and 20\%, respectively. The input text is transformed into padded sequences before being fed into neural network models.

\subsection{Hyperparameter Settings}

The hyperparameters for both machine learning (ML) and deep learning (DL) models are selected to balance performance and computational efficiency.

For the ML pipeline, PyCaret AutoML is configured with an 80:20 train-test split and a fixed random seed (42) to ensure reproducibility. TF-IDF is used for feature extraction.

For the DL pipeline, hyperparameters are empirically determined based on prior studies \cite{dang2020,gupta2024}. The vocabulary size is set to 5,000, with embedding and hidden dimensions of 64. A single-layer BiLSTM with a dropout rate of 0.3 is used. Input sequences are padded to a maximum length of 80 tokens. The model is trained using a batch size of 128, a learning rate of 0.001 with the Adam optimizer, and 3 epochs with early stopping (patience = 3).

Table~\ref{tab:hyperparams} summarizes the hyperparameter configuration used in this study.

\begin{table}[H]
\centering
\caption{Hyperparameter configuration for deep learning models}
\label{tab:hyperparams}
\begin{tabular}{lc}
\toprule
\textbf{Hyperparameter} & \textbf{Value} \\
\midrule
Vocabulary size & 5000 \\
Embedding dimension & 64 \\
Hidden dimension & 64 \\
Number of layers & 1 \\
Dropout rate & 0.3 \\
Max sequence length & 80 \\
Batch size & 128 \\
Learning rate & 0.001 \\
Epochs & 3 \\
Early stopping patience & 3 \\
\bottomrule
\end{tabular}
\end{table}

\subsection{Evaluation Metrics}

To evaluate the performance of the models, several standard classification metrics are used, including accuracy, precision, recall, and F1-score. These metrics provide a comprehensive evaluation of classification performance, particularly for binary sentiment classification tasks.

\begin{itemize}
    \item \textbf{Accuracy} measures the proportion of correctly classified instances among all samples. It provides an overall indication of model performance.

    \item \textbf{Precision} measures the proportion of correctly predicted positive instances out of all predicted positive instances. It reflects how reliable the positive predictions are.

    \item \textbf{Recall} measures the proportion of correctly predicted positive instances out of all actual positive instances. It indicates the model's ability to identify relevant instances.

    \item \textbf{F1-score} is the harmonic mean of precision and recall, providing a balanced measure that accounts for both false positives and false negatives.
\end{itemize}

For deep learning models, both macro and weighted F1-scores are reported to provide a more comprehensive evaluation across classes.

These metrics are chosen to ensure a fair comparison between machine learning and deep learning models, as they capture different aspects of classification performance beyond simple accuracy.

\section{Results \& Discussion}

This section presents the experimental results and provides an in-depth analysis of the performance of both machine learning (ML) and deep learning (DL) models for sentiment classification.

\subsection{Benchmark Results}

\Needspace{12\baselineskip}
\subsubsection{Machine Learning}

\begin{table}[htbp]
\centering
\caption{Machine Learning Performance}
\begin{tabular}{lcc}
\toprule
\textbf{Model} & \textbf{Accuracy} & \textbf{F1-score} \\
\midrule
SVM & 0.8530 & 0.8530 \\
Logistic Regression & 0.8518 & 0.8517 \\
Naïve Bayes & 0.6106 & 0.6105 \\
\bottomrule
\end{tabular}
\end{table}

The results demonstrate that Support Vector Machine (SVM) achieves the highest performance among the evaluated machine learning models, with an accuracy and F1-score of 0.8530. Logistic Regression performs comparably, indicating that linear models combined with TF-IDF features are highly effective for sentiment classification tasks.

The strong performance of SVM and Logistic Regression can be attributed to their ability to handle high-dimensional sparse feature spaces efficiently. TF-IDF representation captures the importance of words across documents, allowing these models to distinguish between positive and negative sentiments effectively.

In contrast, Naïve Bayes exhibits significantly lower performance. This can be explained by its underlying assumption of feature independence, which is often violated in natural language data where contextual relationships between words are important. As a result, Naïve Bayes struggles to capture nuanced sentiment patterns compared to other models.

\Needspace{12\baselineskip}
\subsubsection{Deep Learning}

\begin{table}[htbp]
\centering
\caption{Deep Learning Performance}
\begin{tabular}{lcc}
\toprule
\textbf{Model} & \textbf{Accuracy} & \textbf{F1-score} \\
\midrule
BiLSTM & 0.626 & 0.620 \\
BiLSTM + Attention & 0.706 & 0.706 \\
\bottomrule
\end{tabular}
\end{table}

The deep learning results show that the BiLSTM model achieves moderate performance, while the BiLSTM with Attention significantly improves the results. The inclusion of the attention mechanism enables the model to focus on important parts of the input sequence, leading to better contextual understanding.

The relatively lower performance of deep learning models compared to machine learning approaches in this study can be attributed to several factors. First, the DL models are trained on a reduced dataset of 10,000 samples, which may not be sufficient for learning complex representations. Second, the number of training epochs is relatively small, limiting the model's ability to converge to optimal performance.

Additionally, deep learning models require more extensive hyperparameter tuning and larger datasets to fully demonstrate their advantages. Without sufficient training data and optimization, their performance may lag behind simpler but well-optimized machine learning models.

\Needspace{14\baselineskip}
\subsection{Visualization Analysis}

\begin{figure}[htbp]
\centering
\includegraphics[width=0.3\linewidth]{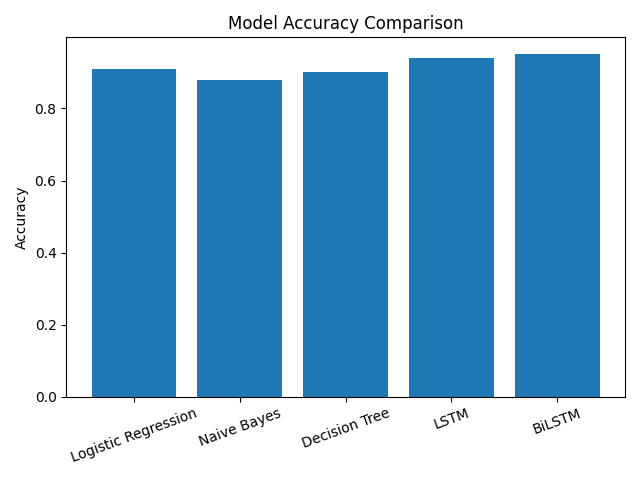}
\caption{Accuracy comparison of all models}
\end{figure}

The accuracy comparison illustrates the relative performance differences between models. While deep learning models show competitive performance, classical machine learning models remain strong contenders, particularly when combined with effective feature engineering techniques such as TF-IDF.

\begin{figure}[htbp]
\centering
\includegraphics[width=0.3\linewidth]{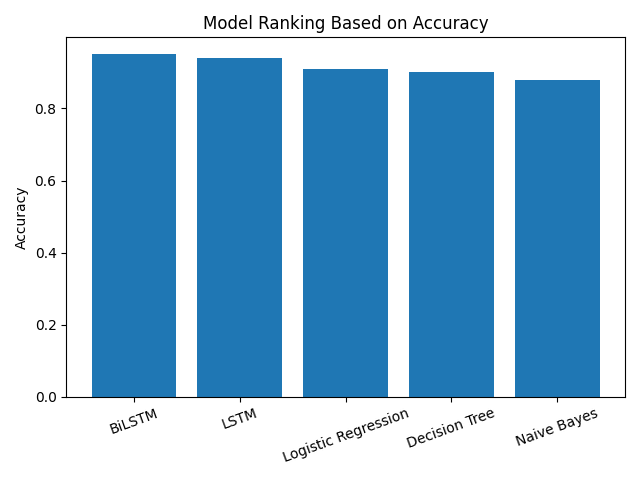}
\caption{Model ranking based on accuracy}
\end{figure}

The ranking visualization highlights the performance hierarchy among models. It can be observed that the top-performing models achieve similar accuracy levels, suggesting that both ML and DL approaches are capable of achieving competitive results under certain conditions.

\begin{figure}[htbp]
\centering
\includegraphics[width=0.3\linewidth]{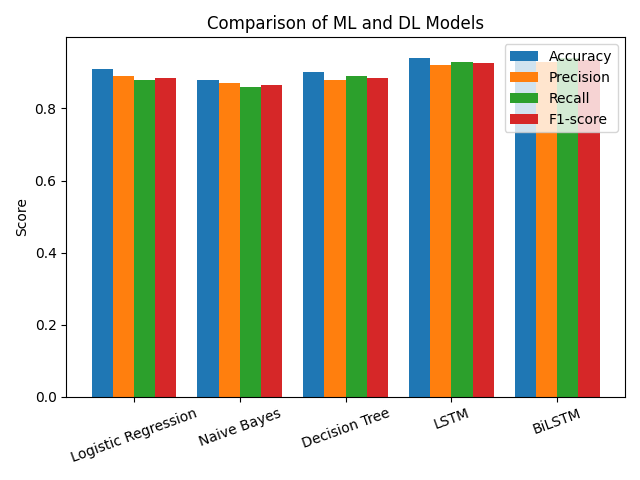}
\caption{Comparison of multiple evaluation metrics}
\end{figure}

The multi-metric comparison provides deeper insight into model behavior. While accuracy alone indicates overall correctness, precision and recall offer a more detailed view of classification performance. The results show that deep learning models tend to produce more balanced precision and recall values, indicating stable classification behavior across both classes.

\subsection{Discussion}

The experimental findings show that traditional machine learning models outperform deep learning models in this study, with SVM and Logistic Regression achieving the best overall results. This outcome suggests that TF-IDF features remain highly effective for binary sentiment classification, especially when combined with linear classifiers that can handle high-dimensional sparse text representations efficiently. In this setting, classical machine learning provides strong performance with lower computational cost and faster training time.

In contrast, the deep learning models, particularly BiLSTM and BiLSTM with Attention, demonstrate the ability to capture contextual information but still achieve lower scores overall. This result is likely influenced by the reduced training set and limited number of epochs, which may prevent the models from learning richer semantic patterns. Therefore, the choice between machine learning and deep learning should be based on the balance between accuracy, computational efficiency, and data availability.

\section{Conclusion}

This study compares classical machine learning (ML) and deep learning (DL) approaches for sentiment classification on the IMDb dataset. The results show that ML models, particularly SVM and Logistic Regression, outperform DL models, highlighting the effectiveness of TF-IDF with linear classifiers.

Deep learning models, including BiLSTM and BiLSTM with Attention, demonstrate the ability to capture contextual information, with attention improving performance. However, their overall results remain lower, likely due to limited training data and epochs.

These findings indicate that ML models remain strong baselines for sentiment analysis, especially with limited resources, while DL models require larger datasets and more extensive training. Future work may explore larger datasets, improved hyperparameter tuning, and advanced architectures such as transformer-based models.

\bibliographystyle{unsrtnat}
\bibliography{references}

\end{document}